\documentclass[10pt,twocolumn,letterpaper]{article}

\usepackage{iccv}
\usepackage{times}
\usepackage{epsfig}
\usepackage{graphicx}
\usepackage{amsmath}
\usepackage{amssymb}
\usepackage{multirow}
\usepackage{multicol}
\usepackage{subcaption}
\usepackage[dvipsnames]{xcolor}
\usepackage{colortbl}
\usepackage{cite}
\usepackage{comment}
\usepackage{verbatim}
\usepackage{enumitem}
\usepackage[acronym]{glossaries}
\usepackage[hang,flushmargin]{footmisc}

\makeglossaries

\usepackage[pagebackref=true,breaklinks=true,letterpaper=true,colorlinks,bookmarks=false]{hyperref}

\iccvfinalcopy 

\def\iccvPaperID{1805}

\definecolor{lightgray}{gray}{0.9}
\definecolor{mediumgray}{gray}{0.8 }

\ificcvfinal\fi
\begin{document}

\title{EPIC-Fusion: Audio-Visual Temporal Binding\\ for Egocentric Action Recognition}

\author{Evangelos Kazakos$^1$~~~~Arsha Nagrani$^2$~~~~Andrew Zisserman$^2$~~~~Dima Damen$^1$\\
$^1$Visual Information Lab, University of Bristol~~~~$^2$Visual Geometry Group, University of Oxford\\
}

\maketitle

\ificcvfinal\thispagestyle{empty}\fi
\begin{abstract}

We focus on multi-modal fusion for egocentric action recognition, 
and propose a novel architecture for multi-modal temporal-binding, i.e. the combination of modalities within a range of temporal offsets. We train the architecture with three modalities -- RGB, Flow and Audio -- and combine them with
mid-level fusion alongside sparse temporal sampling of fused representations. 
In contrast with previous works, modalities are fused before temporal aggregation, with shared modality and fusion weights over time.
Our proposed architecture is trained end-to-end, outperforming individual modalities as well as late-fusion of modalities.

We demonstrate the importance of audio in egocentric vision, on per-class basis, for identifying actions as well as interacting objects.
Our method achieves state of the art results on both the seen and unseen test sets of the largest egocentric dataset: EPIC-Kitchens, on all metrics using the public leaderboard.

\end{abstract}

\section{Introduction}
With the availability of multi-sensor wearable devices (e.g.\ GoPro, Google Glass, Microsoft Hololens, MagicLeap), egocentric audio-video recordings have become popular in many areas such as extreme sports, health monitoring, life logging, and home automation.
As a result, there has been a renewed interest from the computer vision community on collecting large-scale datasets~\cite{Damen_2018_ECCV,sigurdsson2018charadesego} as well as developing new or adapting
existing methods to the first-person point-of-view scenario~\cite{zhou2015temporal,Pirsiavash2012,Lee2012,Ma_2016_CVPR,Damen2014a,Yonetani2016}.

In this work, we explore audio as
a prime modality to provide complementary information to visual
modalities (appearance and motion) in egocentric action recognition. While audio has been explored in video understanding in general ~\cite{ObjectSound,AytarVT16,ArandjelovicZ17,Owens0MFT16,AVEearlyFusion,AytarVT17,VoicesFaces,TextVideo,Senocak_2018_CVPR,gao2019visual-sound} the egocentric domain in particular offers rich sounds resulting from the interactions between hands and objects, as well as the close proximity of the wearable microphone to the undergoing action. Audio is a prime discriminator for some actions (e.g.\ `wash', `fry') as well as objects within actions (e.g.\ `put plate' vs `put bag').
At times, the temporal progression (or change) of sounds can separate visually ambiguous actions (e.g.\ `open tap' vs `close tap'). 
Audio can also capture actions that are out of the wearable camera's field of view, but audible (e.g.\ `eat' can be heard but not seen).
Conversely, other actions are \textit{sound-less} (e.g.\ `wipe hands') and the wearable sensor might capture irrelevant sounds, such as talking or music playing in the background.
The opportunities and challenges of incorporating audio in egocentric action recognition allow us to explore new multi-sensory fusion approaches, particularly related to the potential \textit{temporal asynchrony} between the action's appearance and the discriminative audio signal -- the main focus of our work.

 \begin{figure}[t]
	\centering
	\includegraphics[width=\linewidth]{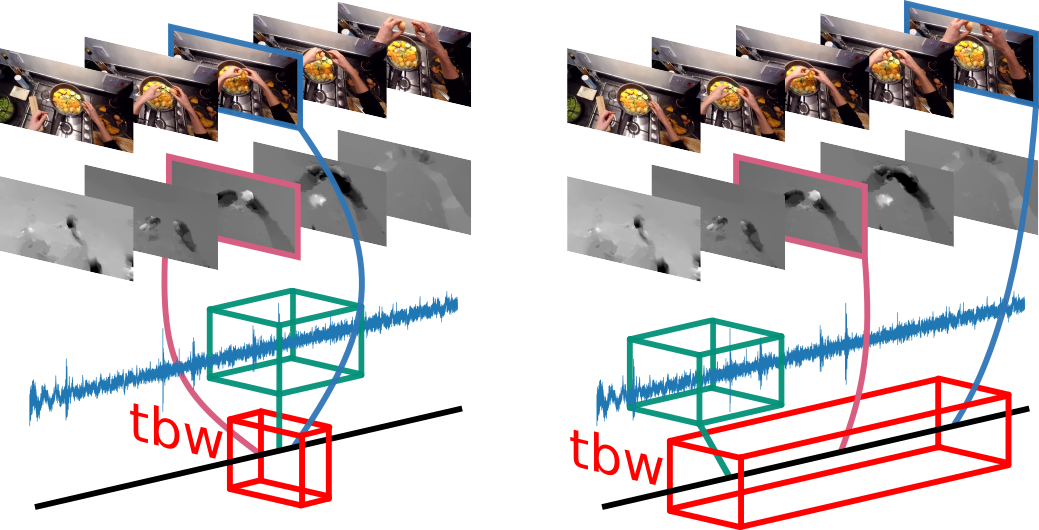}
	\caption{As the width of the temporal binding window increases (left to right), modalities (appearance, motion and audio) are fused with varying temporal shifts.}
	\label{fig:tbw_sampling}
\end{figure}
While several multi-modal fusion architectures exist for action recognition, current approaches perform temporal aggregation {\em within} each modality~\textit{before} modalities are fused \cite{TSN2016ECCV,MiechLS17} or embedded~\cite{TextVideo}. Works that do fuse inputs before temporal aggregation, e.g.~\cite{Feichtenhofer_2016_CVPR}, do so with inputs synchronised across modalities.
In Fig.~\ref{fig:tbw_sampling}, we show an example of `breaking an egg into a pan' from the EPIC-Kitchens dataset. The distinct sound of cracking the egg, the motion of separating the egg and the change in appearance of the egg occur at different frames/temporal positions within the video. Approaches that fuse modalities with synchronised input would thus be limited in their ability to learn such actions.
In this work, we explore fusing inputs within a Temporal Binding Window (TBW) (Fig~\ref{fig:tbw_sampling}), allowing the model 
to train using asynchronous inputs from the various modalities. 
Evidence in neuroscience and behavioural sciences points at the presence of such a TBW in humans~\cite{PARISE201246,WALLACE2014105}. The TBW offers a ``range of temporal offsets within which an individual is able to perceptually bind inputs across sensory modalities''~\cite{Stevenson2013}.
This is triggered by the gap in the biophysical time to process different senses~\cite{Megavand2013}.
Interestingly, the width of the TBW in humans is heavily task-dependant, shorter for simple stimuli such as flashes and beeps and intermediate for complex stimuli such as a hammer hitting a nail~\cite{WALLACE2014105}. 

Combining our explorations into audio for egocentric action recognition, and using a TBW for asynchronous modality fusion, our contributions are summarised as follows. First, an end-to-end trainable mid-level fusion Temporal Binding Network (TBN) is proposed\footnote{ Code at: \url{http://github.com/ekazakos/temporal-binding-network}}. 
Second, we present the first audio-visual fusion attempt in egocentric action recognition.  Third, we achieve state-of-the-art results on the EPIC-Kitchens public leaderboards on both seen and unseen test sets. Our results show (i) the efficacy of audio for egocentric action recognition, (ii) the advantage of mid-level fusion within a TBW over late fusion, and (iii)~the robustness of our model to background or irrelevant sounds. 

\section{Related Work}

We divide the related works into three groups: works that fuse visual modalities (RGB and Flow) for action recognition (AR), works that fuse modalities for egocentric AR in particular, and finally works from the recent surge in interest of audio-visual correspondence and source separation.

\noindent\textbf{Visual Fusion for AR:} By observing the importance of spatial and temporal features for AR, 
two-stream (appearance and motion) fusion has become a standard technique~\cite{NIPS2014_5353, Feichtenhofer_2016_CVPR, TSN2016ECCV}.
\textit{Late fusion}, first proposed by Simonyan and Zisserman~\cite{NIPS2014_5353},  combines the streams' independent predictions. Feichtenhofer \etal~\cite{Feichtenhofer_2016_CVPR} proposed \textit{mid-level fusion} of the spatial and temporal streams, showing optimal results by combining the streams after the last convolutional layer.
In \cite{Carreira_2017_CVPR}, 3D convolution for spatial and motion streams was proposed, followed by late fusion of modalities. 
All these approaches do not model the temporal progression of actions, 
a problem addressed by \cite{TSN2016ECCV}.
Temporal Segment Networks (TSN)~\cite{TSN2016ECCV} perform sparse temporal sampling followed by temporal aggregation (averaging) of softmax scores across samples. 
Each modality is trained independently, 
with late fusion of modalities by averaging their predictions.
Follow-up works focus on pooling for temporal aggregation, still training modalities independently~\cite{Zhou_2018_ECCV,Girdhar_2017_CVPR}.
Modality fusion before temporal aggregation was proposed in~\cite{async_fusion}, where the appearance of the current frame is fused with 5 uniformly sampled motion frames, and vice versa, using two temporal models (LSTM).
While their motivation is similar to ours, their approach focuses on using predefined asynchrony offsets between two modalities. In contrast, we relax this constraint and allow fusion from any random offset within a temporal window, which is more suitable for scaling up to many modalities.

\noindent\textbf{Fusion in Egocentric AR:} 
Late fusion of appearance and motion has been frequently used in egocentric AR~\cite{Damen_2018_ECCV,Song_2016_CVPR_Workshops,Moltisanti2017,Sudhakaran2018}, as well as extended to additional streams
aimed at capturing egocentric cues
\cite{Ma_2016_CVPR,Singh_2016_CVPR,Song_2016_CVPR_Workshops}. 
In \cite{Ma_2016_CVPR}, 
the spatial stream segments hands and detects objects. The streams are trained jointly with a triplet loss on objects, actions and activities, and fused through concatenation. \cite{Singh_2016_CVPR} uses head motion features, hand masks, and saliency maps, which are stacked and 
fed to both a 2D and a 3D ConvNet, and combined by late fusion. 
All previous approaches have relied on small-scale egocentric datasets, and none utilised audio for egocentric AR.

\noindent\textbf{Audio-Visual Learning:} Over the last three years, significant attention has been paid in computer vision to an underutilised and readily available
source of information existing in video: the audio stream
~\cite{ObjectSound,AytarVT16,ArandjelovicZ17,Owens0MFT16,AVEearlyFusion,AytarVT17,VoicesFaces,TextVideo,Senocak_2018_CVPR,gao2019visual-sound}. 
These fall in one of four categories: i) \textit{audio-visual representation learning} \cite{ArandjelovicZ17,AytarVT16,AytarVT17,TextVideo,AVEearlyFusion, Owens0MFT16}, 
ii) \textit{sound-source localisation} \cite{ObjectSound,AVEearlyFusion,Senocak_2018_CVPR},  
iii) \textit{audio-visual source separation}~\cite{AVEearlyFusion, gao2019visual-sound} and (iv) \textit{visual-question answering}~\cite{alamri@DSTC7}.
These approaches attempt fusion \cite{ArandjelovicZ17, AVEearlyFusion} or embedding into a common space~\cite{ObjectSound,AytarVT17,learnablePINs}. 
Several works sample the two modalities with temporal shifts, for learning better synchronous representations~\cite{AVEearlyFusion,Korbar18}. Others
sample within a 1s temporal window, to learn a correspondence between 
the modalities, e.g.\ \cite{ArandjelovicZ17,ObjectSound}.
Of these works, \cite{AVEearlyFusion,Korbar18} note this audio-visual representation learning could be used for AR, by pretraining on the self-supervised task and then fine-tuning for AR. 

 \begin{figure*}[t]
	\centering
	\includegraphics[width=1\textwidth]{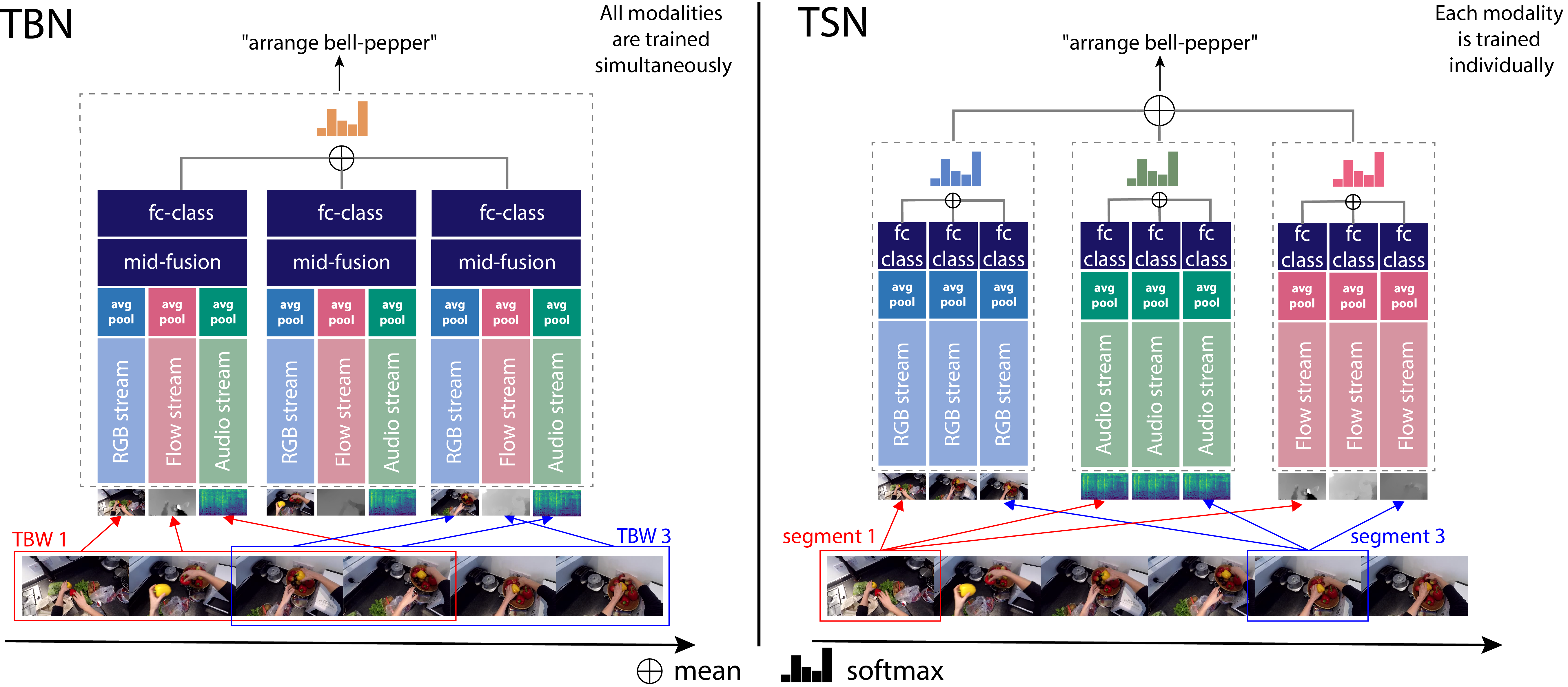}
	\vspace*{-6pt}
	\caption{
	\textbf{Left:} our proposed Temporal Binding Network (TBN). Modalities are sampled within a TBW, and modality-specific weights (same colour) are shared amongst different inputs.
	Modalities are fused with mid-level fusion and trained jointly. Predictions from multiple TBWs, possibly overlapping, are averaged.
	\textbf{Right:} TSN~\cite{TSN2016ECCV} with an additional audio stream performing~\textit{late} fusion. Modalities are trained independently. Note that while in TSN a prediction is made for each modality, TBN produces a single prediction per TBW after fusing all modality representations. Best viewed in colour.}
	\label{fig:tsn}
\end{figure*}

Fusion for AR using three modalities (appearance, motion and audio) has been explored in \cite{WuMM2016}, employing late-fusion of predictions, and \cite{Long_2018_CVPR, AAAI1817054} using attention
to integrate local features into a global representation.
Tested on UCF101, \cite{WuMM2016} shows audio to be the least informative modality for third person action recognition (16\% accuracy for audio compared to 80\% and 78\% for spatial and motion). 
A similar conclusion was made for other third-person datasets (AVA~\cite{Girdhar2018} and Kinetics~\cite{Long_2018_CVPR,AAAI1817054}). 

In this work, we show audio to be a competitive modality for egocentric AR on EPIC-Kitchens, achieving comparable performance to appearance. We also demonstrate that audio-visual modality fusion in egocentric videos improves the recognition performance of both the action and the accompanying object.

\vspace*{-1pt}
\section{The Temporal Binding Network}
\label{sec:tbw}
\vspace*{-3pt}
Our goal is to find the optimal way to fuse multiple modality inputs while modelling temporal progression through sampling. 
We first explain the general notion of temporal binding of multiple modalities in Sec~\ref{sec:method:tbw}, then detail our architecture in Sec~\ref{sec:method:arch}. 

\subsection{Multimodal Temporal Binding}
\label{sec:method:tbw}
Consider a sequence of samples from one modality in a video stream,
$m_i = (m_{i1}, m_{i2}, \cdots, m_{iT/r_i})$
where $T$ is the video's duration and $r_i$ is the modality's framerate (or frequency of sampling).
Input samples are first passed through unimodal feature extraction functions~$f_i$.
To account for varying representation sizes and frame-rates, most multi-modal architectures apply pooling functions $G$ to each modality in the form of average pooling or other temporal pooling functions (e.g. maximum or VLAD~\cite{Jegou2010}), before attempting multimodal fusion. 

Given a pair of modalities $m_1$ and $m_2$, the final class predictions for a video are hence obtained as follows:
\begin{equation}\label{eq:unimodal_aggregation}
    y = h\big(G(f_1(m_1)), G(f_2(m_2))\big)
\end{equation}
where $f_1$ and $f_2$ are unimodal feature extraction functions, $G$ is a temporal aggregation function, $h$ is the multimodal fusion function and $y$ is the output label for the video. 
In such architectures (e.g. TSN \cite{TSN2016ECCV}), modalities are temporally aggregated for a prediction before different modalities are fused; this is typically referred to as `late fusion'. 

Conversely, multimodal fusion can be performed at~\textit{each} time step as in~\cite{Feichtenhofer_2016_CVPR}. One way to do this would be to synchronise modalities and perform a prediction at \textit{each} time-step. For modalities with matching frame rates, synchronised multi-modal samples can be selected as $(m_{1j}, m_{2j})$, and fused according to the following equation:
\begin{equation}\label{eq:sync}
    y = h\big(G(f_{sync}(m_{1j}, m_{2j}))\big)
\end{equation}
where $f_{sync}$ is a multimodal feature extractor that produces a representation for each time step $j$, and $G$ then performs temporal aggregation over all time steps. When frame rates vary, and more importantly so do representation sizes,  only approximate synchronisation can be attempted,
\begin{equation} \label{eq:approximate_sync}
    y = h\big(G(f_{sync}(m_{1j}, m_{2k}))\big) \quad: k = \lceil \frac{j r_2} {r_1} \rceil 
\end{equation}
We refer to this approach as `synchronous fusion' where synchronisation is achieved or approximated.

\textbf{In this work,} however, we propose fusing modalities within temporal windows. Here modalities are fused within a range of temporal offsets, with all offsets constrained to lie within a finite time window, which we henceforth refer to as a temporal binding window (TBW). Formally,
\begin{equation}\label{eq:tbw_fusion}
    y = h\big(G( f_{tbw}(m_{1j}, m_{2k}))\big) \quad: k \in \big[\lceil \frac{jr_2}{r_1}-b\rceil, \lceil \frac{jr_2}{r_1}+b\rceil\big]
\end{equation}
where $f_{tbw}$ is a multimodal feature extractor that combines inputs within a binding window of width $\pm b$.
Interestingly, as the number of modalities increases, say from two to three modalities, the TBW representation allows fusion of modalities each with different temporal offsets, yet within the same binding window $\pm b$:

\vspace*{-8pt}
{\footnotesize{
 \begin{align}\label{eq:multimodal_tbw}
    y = h\big(G(f_{tbw}(m_{1j}, m_{2k}, m_{3l})\big)
    : k \in \big[\lceil \frac{jr_2}{r_1}-b\rceil, \lceil \frac{jr_2}{r_1}+b\rceil\big] \nonumber\\
    \vspace*{-4pt}
    : l \in \big[\lceil \frac{jr_3}{r_1}-b\rceil, \lceil \frac{jr_3}{r_1}+b\rceil\big] 
\end{align}}}
\vspace*{-8pt}

\noindent This formulation hence allows a large number of different inputs combinations to be fused. 
This is different from proposals that fuse inputs over predefined temporal differences (e.g.~\cite{async_fusion}). Sampling within a temporal window allows fusing modalities with various temporal shifts, \textit{up to} the temporal window width~$\pm b$. This: 1) enables straightforward scaling to multiple modalities with different frame rates, 2)~allows training with a variety of temporal shifts, accommodating, say, different speeds of action performance and 3)~provides a natural form of data augmentation.

With the basic concept of a TBW in place, we now describe our proposed audio-visual fusion model, TBN.

\subsection{TBN with Sparse Temporal Sampling}
\label{sec:method:arch}
Our proposed TBN architecture is shown in Fig~\ref{fig:tsn} (left).
First, the action video is divided into $K$ segments of equal width.
Within each segment, we select a random sample of the first modality $\forall k \in K: m_{1k}$.
This ensures the temporal progression of the action is captured by sparse temporal sampling of this modality, as with previous works~\cite{TSN2016ECCV,Zhou_2018_ECCV}, while random sampling within the segment offers further data for training.
The sampled $m_{1k}$ is then used as the centre of a TBW of width $\pm b$.
The other modalities are selected randomly from within each TBW (Eq. \ref{eq:multimodal_tbw}).
In total, the input to our architecture in both training and testing is $K \times M$ samples from $M$ modalities.

\begin{figure}[t]
	\centering
	\includegraphics[width=0.5\textwidth]{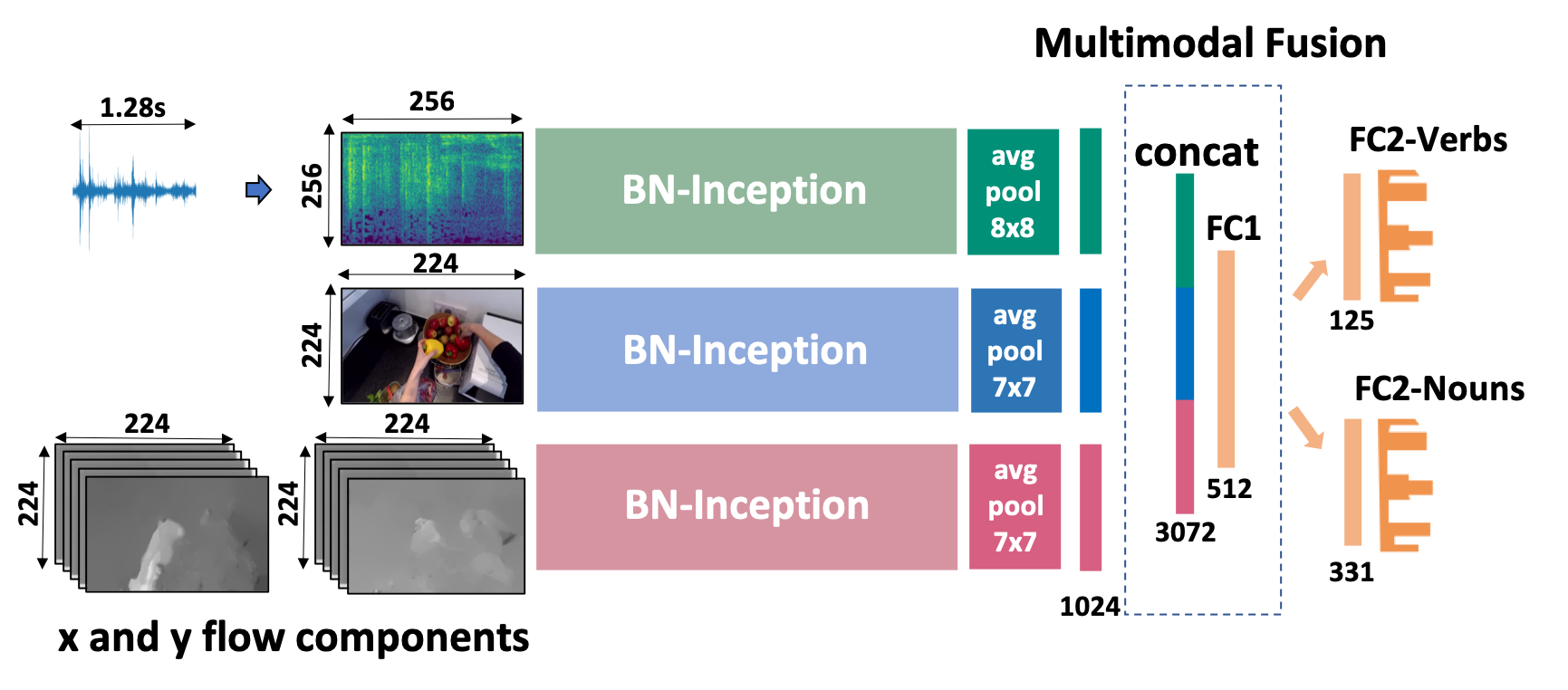}
	\caption{A single TBN block showing architectural details and feature sizes. Outputs from multiple TBN blocks are averaged as shown in Fig.~\ref{fig:tsn}. We model the problem of learning both verbs and nouns as a multi-task learning problem, by adding two output FC layers, one that predicts verbs and the other nouns (as in \cite{Damen_2018_ECCV}). Best viewed in colour.}
	\label{fig:tsn2}
\end{figure}

Within each of the $K$ TBWs, we argue that the complementary information in audio and vision can be better exploited by combining the internal representations
of each modality before temporal aggregation, and hence we propose a \textit{mid-level} fusion. A ConvNet (per modality) extracts \textit{mid-level} features, which are then fused 
through
\textit{concatenating} the modality features and feeding them to a fully-connected layer, making multi-modal predictions per TBW. 
We backpropagate all the way to the inputs of the ConvNets.
Fig~\ref{fig:tsn2} details the proposed TBN block.
The predictions, for each of these unified multimodal representations, are then aggregated for video-level predictions.
In the proposed architecture, we train all modalities simultaneously. 
The convolutional weights for each modality are shared over the $K$ segments. 
Additionally, mid-level fusion weights and class prediction weights are also shared across the segments.

To avoid biasing the fusion towards longer or shorter action lengths, we calculate the window width $b$ relative to the action video length.
Our TBW is thus of variable width, where the width is a function of the length of the action. We note again that $b$ can be set independently of the number of segments $K$, allowing the temporal windows to overlap. This is detailed in Sec.~\ref{sec:implementation}. 

\vspace*{-6pt}
\paragraph{Relation to TSN.} In Fig~\ref{fig:tsn}, we contrast the TBN architecture (left) to an extended version of the TSN architecture (right). The extension is to include the audio modality, since the original TSN only utilises appearance and motion streams. There are two key differences: first, in TSN each modality is temporally aggregated independently (across segments), and the modalities are only combined by late fusion (e.g.\ the RGB scores of each segment are temporally aggregated, and the flow scores of each segment are temporally aggregated, individually). Hence, it is not possible to benefit from combining modalities {\em within} a segment which is the case for TBN. Second, in TSN, each modality is trained independently first after which predictions are combined in inference. 
In the TBN model instead, all modalities are trained simultaneously, and their combination is also learnt.

\section{Experiments}
\noindent \textbf{Dataset:} 
We evaluate the TBN architecture on the largest dataset in egocentric vision: EPIC-Kitchens~\cite{Damen_2018_ECCV},
which contains $39,596$ action segments recorded by $32$ participants performing non-scripted daily activities in their native kitchen environments. In EPIC-Kitchens, an action
is defined as a combination of a \textit{verb} and a \textit{noun}, \eg `cut cheese'. There are in total $125$ verb classes and $331$
noun classes, though these are heavily-imbalanced. The test set is divided in two splits: Seen Kitchens (S1) where sequences from the same environment are in both training, and Unseen Kitchens (S2) where the complete sequences for $4$ participants are held out for testing.
Importantly, EPIC-Kitchens sequences have been captured using a head-mounted Go-Pro with the audio released as part of the dataset. No previous baseline on using audio for this dataset is available.

\subsection{Implementation Details}\label{sec:implementation}
\noindent\textbf{RGB and Flow:} We use the publicly available RGB and computed optical flow with the dataset~\cite{Damen_2018_ECCV}. 

\noindent\textbf{Audio Processing:} We extract $1.28$s of audio, convert it to single-channel, and resample it to 24kHz. We then convert it to a log-spectrogram representation using an STFT of window length $10$ms, hop length $5$ms and
$256$ frequency bands. This results in a 2D spectrogram matrix of size $256\times256$, after which we compute the logarithm. Since many egocentric actions are very short ($<1.28$s),
we extract $1.28$s of audio from the untrimmed video, allowing the audio segment to extend beyond the action boundaries.

\noindent\textbf{Training details:}
We implement our model in PyTorch~\cite{paszke2017automatic}. 
We use Inception with Batch Normalisation (BN-Inception) \cite{pmlr-v37-ioffe15} as a
base architecture, and fuse the modalities after the average pooling layer. 
We chose BN-Inception as it offers a good compromise between performance and model-size, critical
for our proposed TBN that trains all modalities simultaneously, and hence is memory-intensive. Compared to TSN, the three modalities have 10.78M, 10.4M and 10.4M parameters, with only one modality in memory during training. In contrast, 
TBN has 32.64M paramaters.

We train using SGD with momentum \cite{Qian:1999}, a batch size of $128$, a dropout of $0.5$, 
a momentum of $0.9$, and a learning rate of $0.01$. Networks are trained for $80$ epochs, and the learning
rate is decayed by a factor of 10 at epoch $60$. 
We initialise the RGB and the Audio streams from ImageNet. While for the Flow stream, we use stacks of 10 interleaved horizontal and vertical optical flow frames, and use the pre-trained Kinetics \cite{Carreira_2017_CVPR} model, provided by the authors of \cite{TSN2016ECCV}. 

Note that our network is trained end-to-end for all modalities and TBWs. 
We train with $K=3$ segments over the $M=3$ modalities, with $b = T$, allowing the temporal window to be as large as the action segment.
We test using $25$ evenly spaced samples for each modality, as with the TSN basecode for direct comparison.

\subsection{Results}~\label{sec:results}
\vspace*{-12pt}

This section is organised as follows. First, we show and discuss the performance of single modalities,
and compare them with our proposed TBN, with a special focus on the efficacy of the audio stream. 
Second, we compare different mid-level fusion techniques.
And finally, we investigate the effect of the TBW width on both training and testing.

\begin{figure}[t]
\includegraphics[width=1\linewidth]{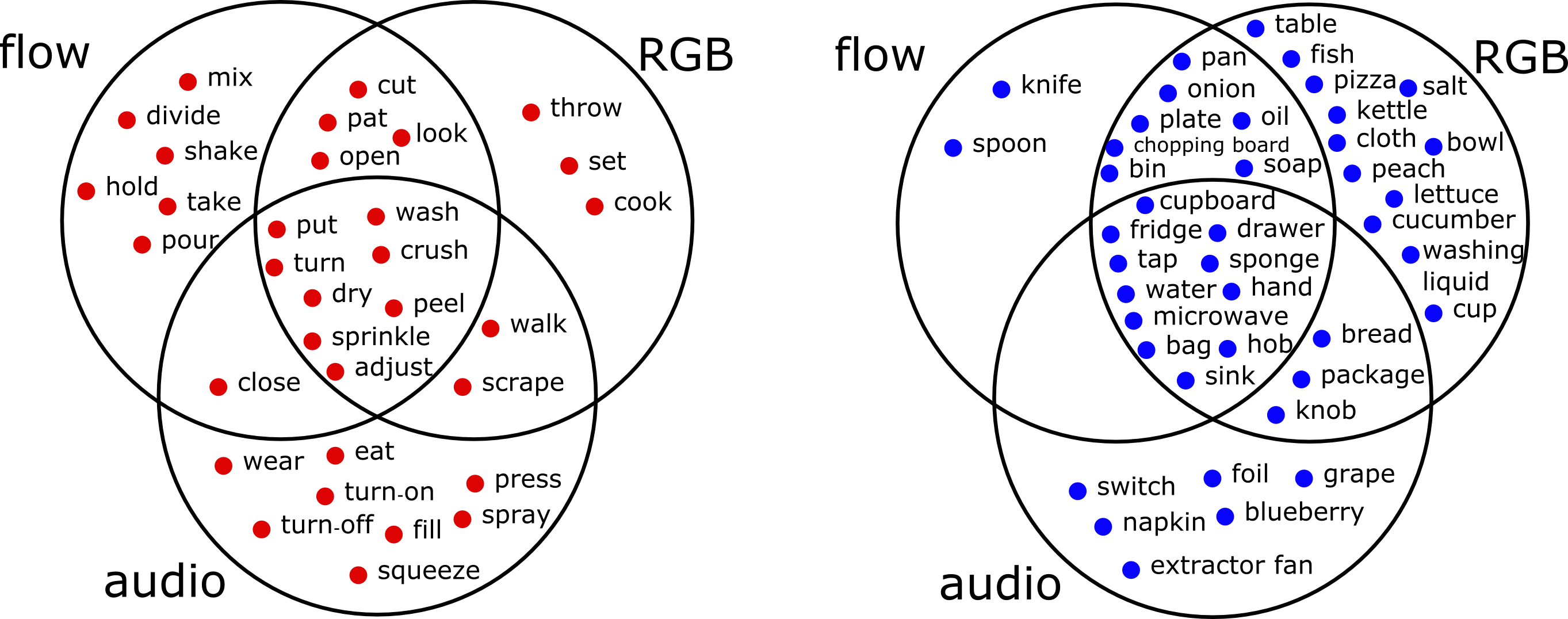}
\caption{Verb (left) and noun (right) classes' performances using single modalities for top-performing 32 verb and 41 noun classes, using single modality accuracy. For each, we consider whether the accuracy is high for Flow, Audio or RGB, or for two or all of these modalities. It can be clearly seen that noun classes can be predicted with high accuracy using RGB alone, whereas for many verbs, Flow and Audio are also important modalities. }
\label{fig:venn}
\end{figure}

\begin{table*}[t]
\vspace{-1mm}
\begin{center}
\resizebox{\textwidth}{!}{
\begin{tabular}{ll|ccc|ccc|ccc|ccc}

                        &        & \multicolumn{3}{c|}{\textbf{Top-1 Accuracy}} & \multicolumn{3}{c|}{\textbf{Top-5 Accuracy}} & \multicolumn{3}{c|}{\textbf{Avg Class Precision}} &\multicolumn{3}{c}{\textbf{Avg Class Recall}} \\\cline{3-14}
                        &        & VERB      & NOUN      & ACTION     & VERB      & NOUN      & ACTION     & VERB   & NOUN  & ACTION & VERB   & NOUN  & ACTION \\\hline
\multirow{5}{*}{\rotatebox{90}{\textbf{S1}}}
                        & RGB & 45.68 & 36.80 & 19.86 & 85.56& 64.19 & 41.89 & 61.64 & 34.32 & 09.96 & 23.81 & 31.62 & 08.81 \\
                        & Flow & 55.65 & 31.17 & 20.10 & 85.99 & 56.00 & 39.30 & 48.83 & 26.84 & 09.02 & 27.58 & 24.15 & 07.89 \\
                        & Audio & 43.56  & 22.35  & 14.21  & 79.66  & 43.68  & 27.82  & 32.28  & 19.10  & 07.27  & 25.33  & 18.16  & 06.17 \\
                        & TBN (RGB+Flow) & 60.87 & 42.93 & 30.31 & 89.68 & 68.63 & 51.81 & \textbf{61.93} & 39.68 & 18.11 & 39.99 & 38.37 & 16.90\\
                        & TBN (All) & \textbf{64.75} & \textbf{46.03} & \textbf{34.80} & \textbf{90.70} & \textbf{71.34} & \textbf{56.65} & 55.67 & \textbf{43.65} & \textbf{22.07} & \textbf{45.55} & \textbf{42.30} & \textbf{21.31}\\
                        \hline
\multirow{5}{*}{\rotatebox{90}{\textbf{S2}}}
                        & RGB & 34.89 & 21.82 & 10.11 & 74.56 & 45.34 & 25.33 & 19.48 & 14.67 & 04.77 & 11.22 & 17.24 & 05.67 \\
                        & Flow & 48.21 & 22.98 & 14.48 & 77.85 & 45.55 & 29.33 & 23.00 & 13.29 & 05.63 & 19.61 & 16.09 & 07.61 \\
                        & Audio & 35.43  & 11.98  & 06.45  & 69.20  & 29.49  & 16.18  & 22.46  & 09.41  & 04.59  & 18.02  & 09.79  & 04.19\\
                        & TBN (RGB+Flow) & 49.61 & 25.68 & 16.80 & 78.36 & 50.94 & 32.61 & 30.54 & 20.56 & 09.89 & 21.90 & 20.62 & 11.21\\
                        & TBN (All) & \textbf{52.69} & \textbf{27.86} & \textbf{19.06} & \textbf{79.93} & \textbf{53.78} & \textbf{36.54} & \textbf{31.44} & \textbf{21.48} & \textbf{12.00} & \textbf{28.21} & \textbf{23.53} & \textbf{12.69} \\

\end{tabular}}
\caption{Comparison of our fusion method to single modality performance. For both splits, the fusion outperforms single modalities. For the seen split, the RGB and Flow modalities perform comparatively, whereas for the unseen split the Flow modality outperforms RGB by a large margin. Audio is comparable to RGB on top-1 verb accuracy for both splits.} 
\label{tab:single}
\end{center}
\end{table*}

\noindent\textbf{Single-modality vs multimodal fusion performance: }We examine the overall performance of each modality individually in Table~\ref{tab:single}. Although it is clear that RGB and optical flow are stronger modalities than audio, an interesting find is that
audio performs comparably to RGB on some of the metrics (e.g. top-1 verb accuracy), signifying 
the relevance of audio on recognising egocentric actions. While
as expected optical flow outperforms RGB in \textbf{S2}, interestingly for \textbf{S1}, the RGB and Flow modalities perform comparatively, and in some cases
RGB performs better. 
This matches the expectation that Flow is more invariant to the environment.

To obtain a better analysis of how these modalities perform, we examine the accuracy of ~\textit{individual} verb and noun classes on \textbf{S1}, using single modalities.
Fig~\ref{fig:venn} plots top-performing verb and noun classes, into a Venn diagram. For each class, we consider the accuracy of individual modalities. If all modalities perform comparably (within 0.15), we plot that class in the intersection of the three circles. On the other hand, if one modality is clearly better than the others (more than 0.15), we plot the class in the outer part of the modality's circle. For example, for the verb `close', we have per-class accuracy of 0.23, 0.47 and 0.42 for RGB, Flow and Audio respectively. We thus note that this class performs best for two modalities: Flow and Audio, and plot it in the intersection of these two circles.

From this plot, many verb and noun classes perform comparably for all modalities (e.g.\ `wash', `peel' and `fridge', `sponge'). This suggests all three modalities contain useful information for these tasks. A distinctive difference, however, is observed in the importance of individual modalities for verbs and nouns. Verb classes are strongly related to the temporal progression of actions, making Flow more important for verbs than nouns. Conversely, noun classes can be predicted with high accuracy using RGB alone. Audio, on the other hand, is important for both nouns and verbs, particularly for some verbs such as `turn-on', and `spray'. For nouns, Audio tends to perform better for objects with distinctive sounds (e.g.\ `switch', `extractor fan') and materials that sound when manipulated (e.g. `foil').

\begin{figure}[t]
	\centering
	\includegraphics[width=0.49\linewidth]{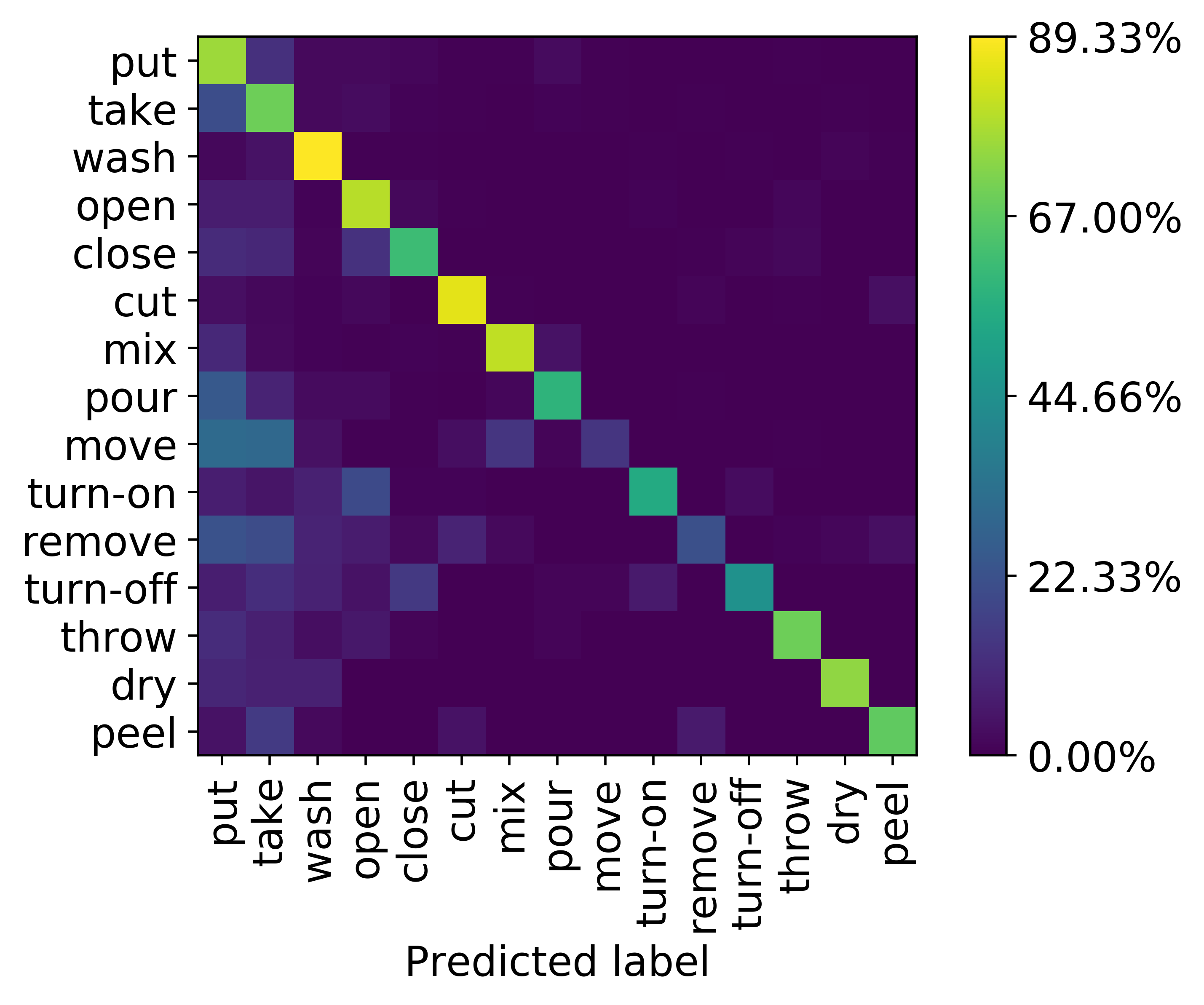}
	\includegraphics[width=0.49\linewidth]{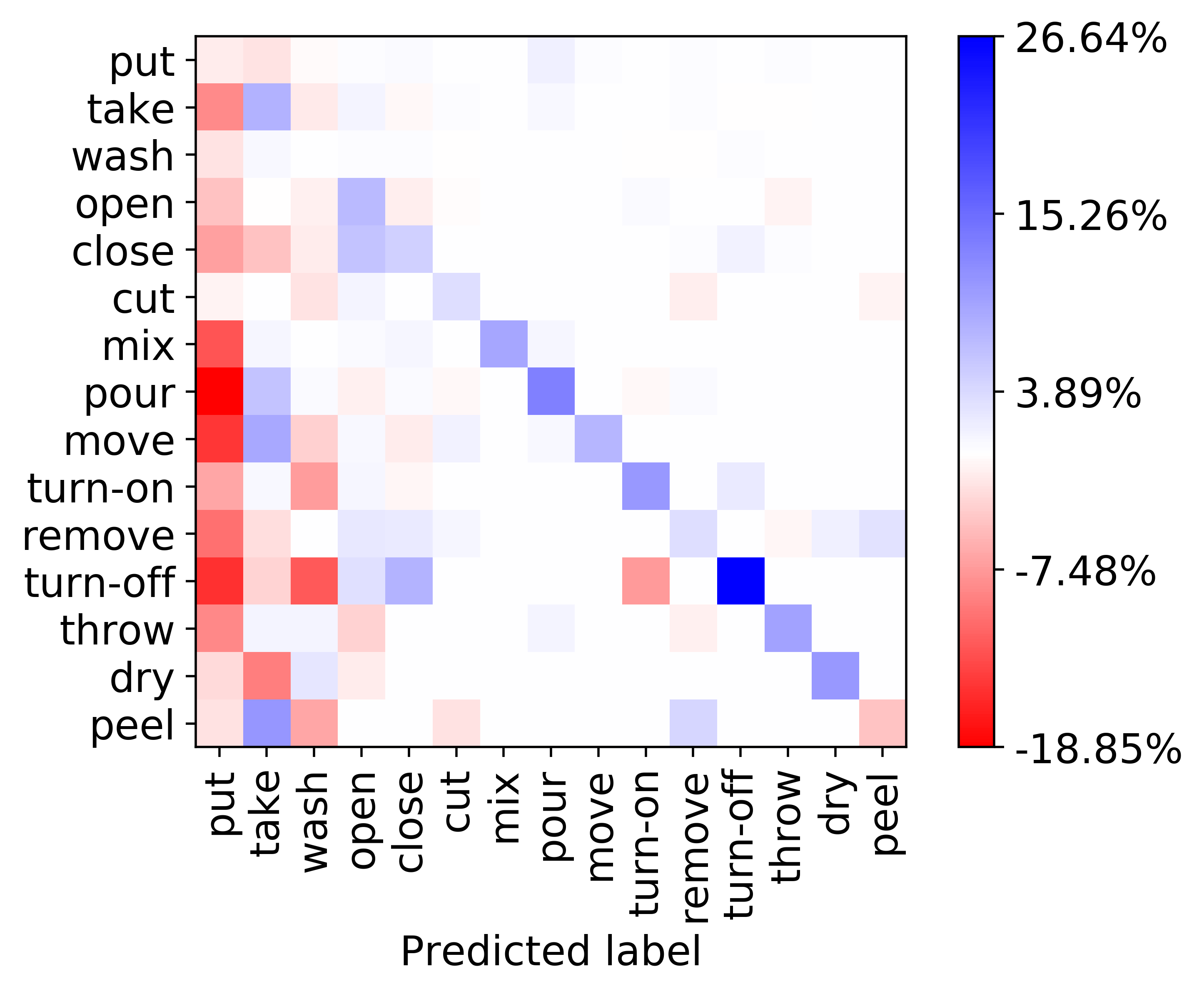} 
	\caption{Confusion matrix for the largest-15 verb classes, with audio (left), as well as the difference to the confusion matrix without audio (right).}
	\label{fig:conf_mat}
\end{figure}

\begin{figure*}[t]
	\centering
	\includegraphics[width=1\textwidth]{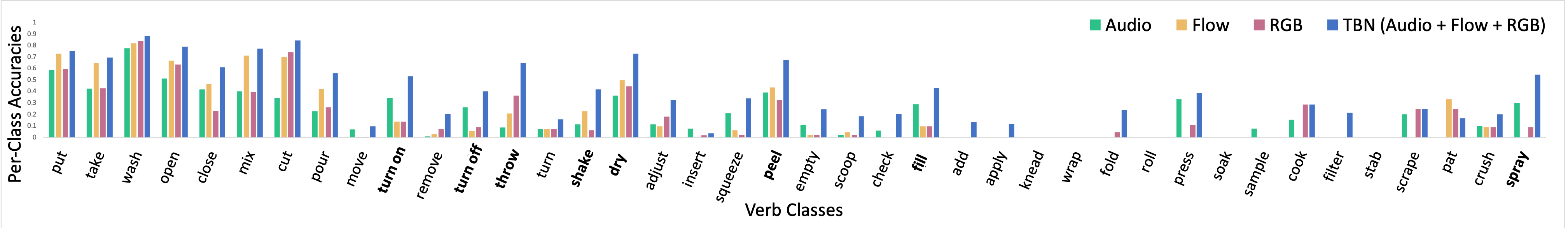}	\includegraphics[width=1\textwidth]{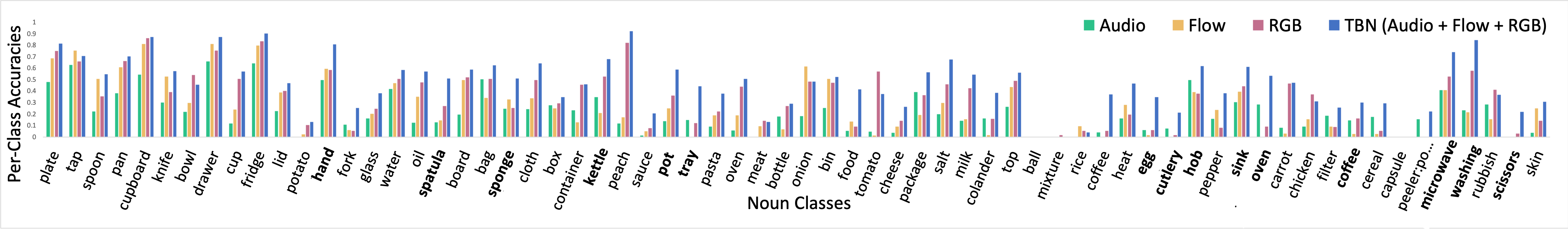}
	\caption{Per-class accuracies for the \textbf{S1} test set for verbs (top) and nouns (bottom) for fusion and single modalities. We  select verb classes with more than 10 samples, and noun classes with more than 30 samples. The classes are presented in the order of number of samples per class, from left to right. For most classes the fusion method provides significant performance gains over single modality classification (largest improvements shown in bold).
	Best viewed in colour.}
	\label{fig:perclass}
\end{figure*}

In Table.~\ref{tab:single}, we compare single modality performance to the performance over the three modalities. Single modalities are trained as in TSN, as TBN is designed to bind multiple modalities. We find that the fusion method outperforms single modalities, and that audio is a significantly informative modality across the board. 
Per-class accuracies, for individual modalities as well as for TBN trained on all three modalities, can be seen in Figure~\ref{fig:perclass}. The advantage of the fusion method is more pronounced for verbs (where we expect motion and audio to be more informative) than nouns, and more for particular noun classes than others, such as `pot', `kettle', `microwave', and particular verb classes eg. `spray' (fusion 0.54, RGB 0.09, Flow 0, Audio 0.3). This suggests that the mixture of complementary and redundant information captured in a video is highly dependant on the action itself, yielding the fusion method to be more useful for some classes than for others. We also note that the fusion method helps to significantly boost the performance of the tail classes (Fig.~\ref{fig:perclass}, right and table in Appendix~\ref{sec:perclass}), where individual modality performance tends to suffer. 

\begin{table}[t]
\vspace{-1mm}
\begin{center}
\resizebox{1\linewidth}{!}{
\begin{tabular}{ll|ccc|ccc}

                        &        & \multicolumn{3}{c|}{\textbf{All}} & \multicolumn{3}{c}{\textbf{RGB+Flow}}\\\cline{3-8}
                        & TBN        & VERB      & NOUN      & ACTION     & VERB      & NOUN      & ACTION \\\hline
\multirow{2}{*}{\rotatebox{90}{\textbf{S1}}}
                        & irrelevant & 61.37 & 46.46 & 32.63 & 57.28 & 42.55 & 27.73\\
                        & rest & 65.28 & 45.97 & 35.14 & 61.44 & 42.99 & 30.72\\
                        \hline
\multirow{2}{*}{\rotatebox{90}{\textbf{S2}}}
                        & irrelevant & 47.32 & 23.36 & 15.30 & 44.41 & 20.45 & 12.39\\
                        & rest & 57.21 & 31.66 & 22.22 & 54.00 & 30.09 & 20.52\\
\end{tabular}}
\caption{Comparing top-1 accuracy of All modalities (left) to RGB+Flow (right). Actions are split in segments with `irrelevant'  background sounds, and the `rest' of the test set.} 
\label{tab:irrelevant}
\end{center}
\end{table}

\begin{table*}[t]
\vspace{-1mm}
\begin{center}
\resizebox{\textwidth}{!}{
\begin{tabular}{ll|ccc|ccc|ccc|ccc}

                        &        & \multicolumn{3}{c|}{\textbf{Top-1 Accuracy}} & \multicolumn{3}{c|}{\textbf{Top-5 Accuracy}} & \multicolumn{3}{c|}{\textbf{Avg Class Precision}} &\multicolumn{3}{c}{\textbf{Avg Class Recall}} \\\cline{3-14}
                        &        & VERB      & NOUN      & ACTION     & VERB      & NOUN      & ACTION     & VERB   & NOUN  & ACTION & VERB   & NOUN  & ACTION \\\hline
\multirow{3}{*}{\rotatebox{90}{\textbf{S1}}}
                        & Concatenation & \textbf{64.75} & \textbf{46.03} & \textbf{34.80} & \textbf{90.70} & \textbf{71.34} & \textbf{56.65} & 55.67 & \textbf{43.65} & \textbf{22.07} & 45.55 & \textbf{42.30} & \textbf{21.31}\\
                        & Context gating~\cite{MiechLS17} & 63.77 & 44.33 & 33.47 & 90.04 & 69.09 & 54.10 & \textbf{57.31} & 42.20 & 21.72 & \textbf{45.63} & 41.53 & 20.20\\
                        & Gating fusion \cite{gated_multimodal} & 61.52 & 43.54 & 31.61 & 89.54 & 68.42 & 52.57 & 52.07 & 39.62 & 18.39 & 42.55 & 39.77 & 18.66\\
                        \hline
\multirow{3}{*}{\rotatebox{90}{\textbf{S2}}}
                        & Concatenation & \textbf{52.69} & \textbf{27.86} & 19.06 & \textbf{79.93} & \textbf{53.78} & \textbf{36.54} & \textbf{31.44} & 21.48 & 12.00 & \textbf{28.21} & \textbf{23.53} & 12.69\\
                        & Context gating~\cite{MiechLS17} & 52.65 & 27.35 & \textbf{19.16} & 79.25 & 52.00 & 36.40 & 30.82 & \textbf{23.16} & 11.72 & 23.39 & 25.03 & 12.58\\
                        & Gating fusion \cite{gated_multimodal} & 50.16 & 27.25 & 18.41 & 78.80 & 50.84 & 34.04 & 28.42 & 22.42 & \textbf{12.34} & 23.92 & 24.15 & \textbf{13.14}\\
                        
\end{tabular}}
\caption{Comparison of mid-level fusion techniques for the TBN architecture.}
\label{tab:fusion}
\end{center}
\end{table*}

\noindent \textbf{Efficacy of audio:} We train
TBN only with the visual modalities (RGB+Flow) and the results can be seen in Table~\ref{tab:single}. An increase of  $5\%$ (S1) and $4\%$ (S2) in
top-5 action recognition accuracy with the addition of audio demonstrates
the importance of audio for egocentric action recognition. 
Fig~\ref{fig:conf_mat} shows the confusion matrix with the utilisation of audio for
the largest-15 verb classes (in \textbf{S1}). Studying the difference (Fig~\ref{fig:conf_mat}~right) clearly demonstrates an increase~(blue) in confidence along the diagonal, and a decrease~(red) in confusion
elsewhere.

\begin{table*}[t]
\vspace{-1mm}
\begin{center}
\resizebox{\textwidth}{!}{
\begin{tabular}{ll|ccc|ccc|ccc|ccc}

                        &        & \multicolumn{3}{c|}{\textbf{Top-1 Accuracy}} & \multicolumn{3}{c|}{\textbf{Top-5 Accuracy}} & \multicolumn{3}{c|}{\textbf{Avg Class Precision}} &\multicolumn{3}{c}{\textbf{Avg Class Recall}} \\\cline{3-14}
                        &        & VERB      & NOUN      & ACTION     & VERB      & NOUN      & ACTION     & VERB   & NOUN  & ACTION & VERB   & NOUN  & ACTION \\\hline
\multirow{5}{*}{\rotatebox{90}{\textbf{S1}}}
                        & Attention Clusters~\cite{Long_2018_CVPR} & 40.39 & 19.37 & 11.09 & 78.13 & 41.73 & 24.36 & 21.17 & 09.65 & 02.50 & 14.89 & 11.50 & 03.41\\
                        & \cite{Damen_2018_ECCV} (from leaderboard)  &48.23 &36.71 &20.54 &84.09 &62.32 &39.79 &47.26 &35.42 &11.57&22.33 &30.53& 09.78\\
                        & Ours (TSN \cite{TSN2016ECCV} w. Audio) & 55.49 & 36.27 & 23.95 & 87.04 & 64.17 & 44.26 & 53.85 & 30.94 & 13.55 & 30.60 & 29.82 & 11.11 \\
                        & \textbf{Ours (TBN, Single Model)} & 64.75 & 46.03 & 34.80 & 90.70 & 71.34 & 56.65 & 55.67 & 43.65 & 22.07 & 45.55 & 42.30 & 21.31\\ 
                        & \textbf{Ours (TBN, Ensemble)} & \textbf{66.10} & \textbf{47.89} & \textbf{36.66} & \textbf{91.28} & \textbf{72.80} & \textbf{58.62} & \textbf{60.74} & \textbf{44.90} & \textbf{24.02} & \textbf{46.82} & \textbf{43.89} & \textbf{22.92}\\
                        \hline
\multirow{5}{*}{\rotatebox{90}{\textbf{S2}}}
                        & Attention Clusters~\cite{Long_2018_CVPR} & 32.37 & 11.95 & 05.60 & 69.89 & 31.82 & 15.74 & 17.21 & 03.86 & 01.84 & 11.59 & 07.94 & 02.64\\
                        & \cite{Damen_2018_ECCV} (from leaderboard)  &39.40 &22.70 &10.89 &74.29 &45.72 &25.26 &22.54 &15.33 &06.21 &13.06 &17.52 &06.49\\
                        & Ours (TSN \cite{TSN2016ECCV} w. Audio) & 46.61 & 22.50 & 13.05 & 78.19 & 48.59 & 29.13 & 28.92 & 15.48 & 06.47 & 21.58 & 16.61 & 07.55\\
                        & \textbf{Ours (TBN, Single Model)} & 52.69 & 27.86 & 19.06 & 79.93 & 53.78 & 36.54 & 31.44 & 21.48 & \textbf{12.00} & \textbf{28.21} & 23.53 & 12.69\\
                        
                        & \textbf{Ours (TBN, Ensemble)} & \textbf{54.46} & \textbf{30.39} & \textbf{20.97} & \textbf{81.23} & \textbf{55.69} & \textbf{39.40} & \textbf{32.57} & \textbf{21.68} & 10.96 & 27.60 & \textbf{25.58} & \textbf{13.31}\\

\end{tabular}}
\caption{Results on the EPIC-Kitchens for seen (S1) and unseen (S2) test splits. At the time of submission, our method outperformed all previous methods on all metrics, and in particular by 11\%, 5\% and 4\% on top-1 verb, noun and action classification on S1. Our method achieved second ranking in the 2019 challenge. Screenshots of the leaderboard at submission and challenge conclusion are in the supplementary material.}
\label{tab:resultsF}
\end{center}
\end{table*}

\noindent \textbf{Audio with irrelevant sounds:} In the recorded videos for EPIC-Kitchens, background sounds irrelevant to the observed actions have been captured by the wearable sensor.
These include music or TV playing in the background, ongoing washing machine, coffee machine or frying sounds while actions take place. 
To quantify the effect of these sounds, we annotated the audio in the test set, and report that $14\%$ of all action segments in S1, and $46\%$ of all action segments in S2 contain other audio sources. We refer to these as actions containing `irrelevant' sounds, and independently report the results in Table~\ref{tab:irrelevant}. 
The table shows that the model's accuracy increases consistently when audio is incorporated, even for the `irrelevant' segments.
Both models (All and RGB+Flow) show a drop in performance for `irrelevant' S2 (comparing to `rest'), validating that irrelevant sounds are not the source of confusion, but that this set of action segments is more challenging even in the visual modalities.
This demonstrates the robustness of our network to noisy and unconstrained audio sources.

\noindent\textbf{Comparison of fusion strategies: }As Fig~\ref{fig:tsn} indicates, TBN performs mid-level fusion on the modalities within the binding window.
Here we describe three alternative mid-level fusion strategies, and then compare their performances.

\noindent \textbf{(i)} {Concatenation}, where the feature maps of each modality are concatenated, and a fully-connected layer is used to model the cross-modal relations.
\begin{equation}
\label{eq:concat}
    f_{tbw}^{concat} = \phi(W[m_{1j},m_{2k},m_{3l}] + b)
\end{equation}
where $\phi$ is a non-linear activation function.  When used within TBWs, shared weights $f_{tbw}$ are to be learnt between modalities within a range of temporal shifts.

\noindent \textbf{(ii)} \textit{Context gating} was used in \cite{MiechLS17}, aiming to recalibrate the strength of the activations of different units with a self-gating mechanism:
\begin{eqnarray}\label{eq:context_gating}
f_{tbw}^{context} &= &\sigma(Wh+b_z) \circ h      
\end{eqnarray}
where $\circ$ is element-wise multiplication. We apply
context gating on top of our multi-modal fusion with concatenation, so $h$ in \eqref{eq:context_gating} is equivalent to \eqref{eq:concat}. 

\begin{figure}[t]
\includegraphics[width=0.48\linewidth]{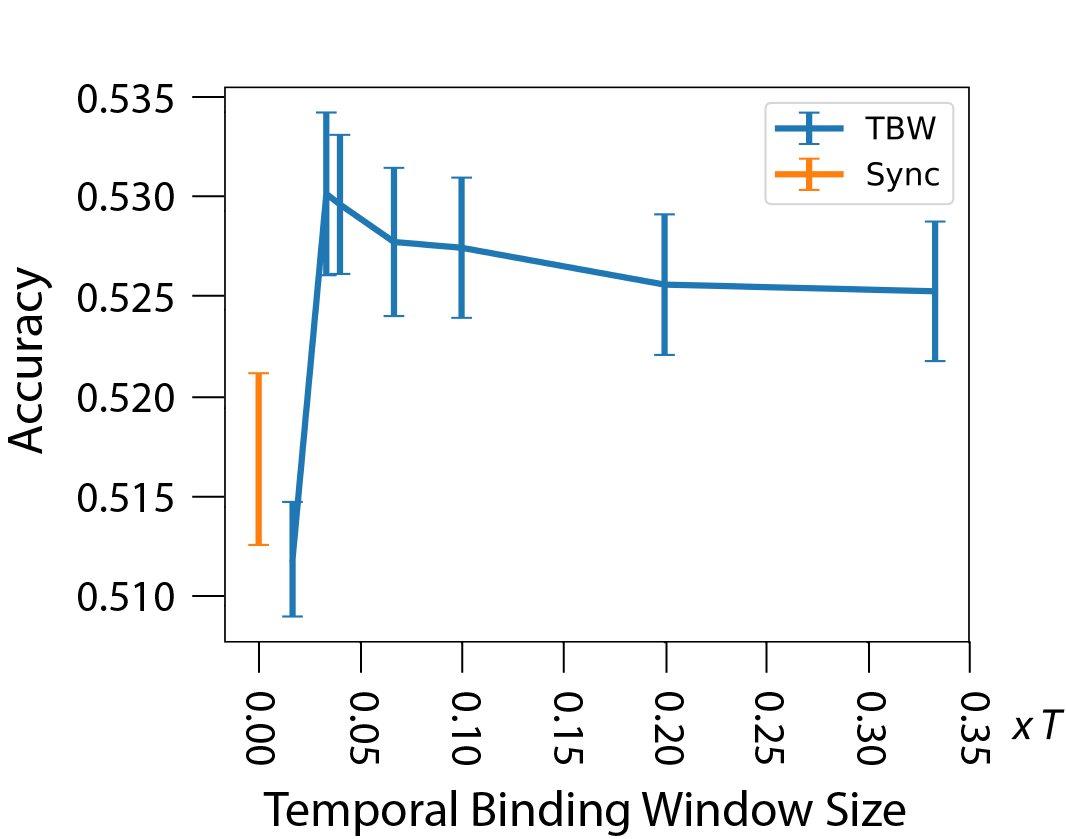} \includegraphics[width=0.48\linewidth]{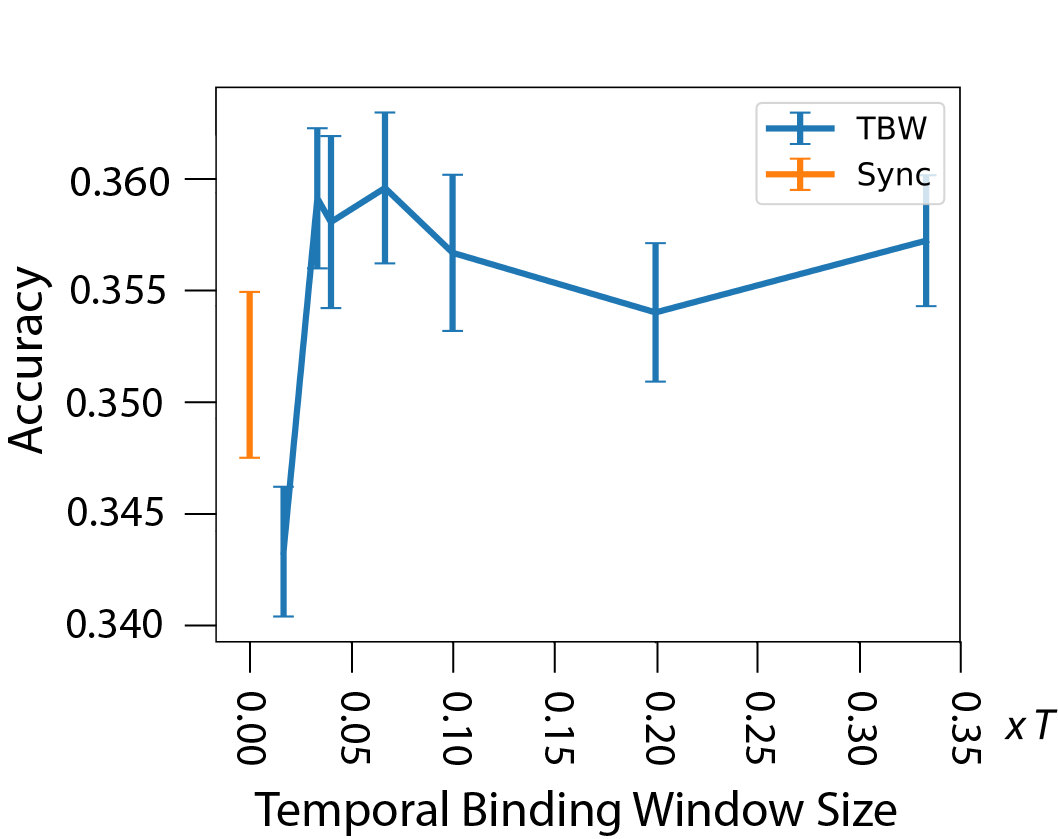}
\caption{Effect of TBW width for verbs (left) and nouns (right) in the \textbf{S1} test set.}
\label{fig:tbw_analysis}
\end{figure}

\noindent \textbf{(iii)} \textit{Gating fusion} was introduced in \cite{gated_multimodal}, where a gate neuron takes as input the features from
all modalities to learn the importance of one modality w.r.t. all modalities. 
\begin{eqnarray}
    h_i&=&\phi(W_im_{ij}+b_i) \quad \forall i \text{     }  \label{eq:h1}\\
    z_i&=&\sigma(W_{zi}[m_{1j},m_{2k},m_{3l}]+b_{z_i}) \quad \forall i \text{       }\\
    f_{tbw}^{gating}&=&z_1\circ h_1 + z_2\circ h_2 + z_3\circ h_3\label{eq:unconstrained}\text{,}
\end{eqnarray}
\noindent In Table~\ref{tab:fusion}, we compare the various fusion strategies. We find that the simplest method, concatenation (Eq.~\ref{eq:concat}) generally outperforms 
more complex fusion approaches. 
We believe this shows modality binding within a temporal binding window to be robust to the mid-level fusion method.

\noindent\textbf{The effect of TBW width:}
Here, we investigate the effect of the TBW width in training and testing.
We varied TBW width in training with $b \in \{\frac{T}{6}, \frac{T}{3}, T\}$, by training three TBN models for each respective window width. We noted little difference in performance. As changing $b$ in training is expensive and performance is subject to the particular optimisation run, we opt for a more conclusive test by focusing on varying $b$ in testing for a single model.

In testing, we vary $b \in \{\frac{T}{60}, \frac{T}{30}, \frac{T}{25}, \frac{T}{15}, \frac{T}{10}, \frac{T}{5}, \frac{T}{3}\}$. This corresponds, in average, to varying the width of TBW on the \textbf{S1} test set between 60ms and 1200ms.
We additionally run with synchrony $b \sim 0$. 
In each case we sample a \textit{single TBW}, to solely assess the effect of the window size. We repeat this experiment for 100 runs and report mean and standard deviation in Fig.~\ref{fig:tbw_analysis}, where we compare results for verb and noun classes separately. The figure shows that 
best performance is achieved for $b \in \left[\frac{T}{30}, \frac{T}{20}\right]$, that is on average $b \in \left[120ms\pm190ms, 180ms\pm285ms\right]$. TBWs of smaller width show a clear drop in performance, with synchrony comparable to $b = \frac{T}{60}$. 
Note that the `Sync' baseline provides only approximate synchronisation of modalities, as modalities have different sampling rates (RGB 60fps, flow 30fps, audio 24000kHz).
The model shows a degree of robustness for larger TBWs. 

Note that in Fig.~\ref{fig:tbw_analysis}, we compare widths on a single temporal window in testing. When we temporally aggregate multiple TBWs, the effect of the TBW width is smoothed, and the model becomes robust to TBW widths. 

\noindent \textbf{Comparison with the state-of-the-art: }
We compare our work to the baseline results reported in \cite{Damen_2018_ECCV} in Table~\ref{tab:resultsF} on all metrics.
First we show that a late fusion with an additional audio stream, outperforms the baseline on top-1 verb accuracy by 7\% on S1 and also 7\% on S2. 
Second, we show that our TBN single model, improves these results even further (9\%, 10\% and 11\% on top-1 verb, noun and action accuracy on S1, and 6\%, 5\% and 6\% on S2 respectively). 
Finally we report results of an Ensemble of five TBNs, where each one is trained with a different TBW width. The ensemble shows additional improvement of up to 3\% on top-1 metrics.

We compare TBN with Attention Clusters~\cite{Long_2018_CVPR}, a previous effort to utilise RGB, Flow, and Audio for action recognition, using \textit{pre-extracted features}.
We use the authors available implementation, and fine-tuned features (TSN, BN-Inception), from the global avg pooling layer~(1024D), to
provide a fair comparison to TBN, and follow the implementation choices from~\cite{Long_2018_CVPR}.
The method from~\cite{Long_2018_CVPR} performs significantly worse than the baseline, as pre-extracted video features are used to learn attention weights. 

At the time of submission, our TBN Ensemble results demonstrated an overall improvement over all state-of-the-art, published or anonymous, by 11\% on top-1 verb for both S1 and S2. Our method was also ranked 2nd in the 2019 EPIC-Kitchens Action Recognition challenge. Details of the public leaderboard are provided in Appendix~\ref{sec:leaderboard}.

\section{Conclusion}
We have shown that the TBN architecture is able to flexibly combine the RGB, Flow and Audio modalities to achieve an across the board performance improvement, compared to individual modalities. In particular, we have demonstrated how audio is complementary to appearance and motion for a number of classes; and the pre-eminence of appearance for noun (rather than verb) classes.
The performance of TBN significantly exceeds TSN trained on the same data; and provides state-of-the-art results on the public EPIC-Kitchens leaderboard.

Further avenues for exploration include a model that learns to adjust TBWs over time, as well as implementing class-specific temporal binding windows. 

\vspace{2pt}
\noindent \textbf{Acknowledgements} \hspace{4pt} Research supported by EPSRC LOCATE (EP/N033779/1), GLANCE (EP/N013964/1) \& Seebibyte (EP/M013774/1). EK is funded by EPSRC Doctoral Training Partnership, and AN by a Google PhD Fellowship.

{\small
\bibliographystyle{ieee_fullname}
\bibliography{egbib}
}

\clearpage
\appendix
\section*{Appendices}

This additional material includes a description of the qualitative examples in the supplementary video in App.~\ref{sec:qual}. This is followed by the leaderboard results  in App~\ref{sec:leaderboard}. We further analyse per-class results with and without audio in App.~\ref{sec:perclass}.

\section{Qualitative results}
\label{sec:qual}

We show selected qualitative results on a held-out validation set, from the publicly available training videos.
We hold-out 14 (untrimmed) videos from the training set, for qualitative examples.  
Video can be watched at \url{https://www.youtube.com/watch?v=VzoaKsDvv1o}. For each, we show the ground truth, and the predictions of individual modalities (RGB, Flow, Audio) compared with our TBN~(Single Model).

\section{Action Recognition Challenge - Public Leaderboard Results}
\label{sec:leaderboard}

\begin{figure}[t]
 \includegraphics[width=\linewidth]{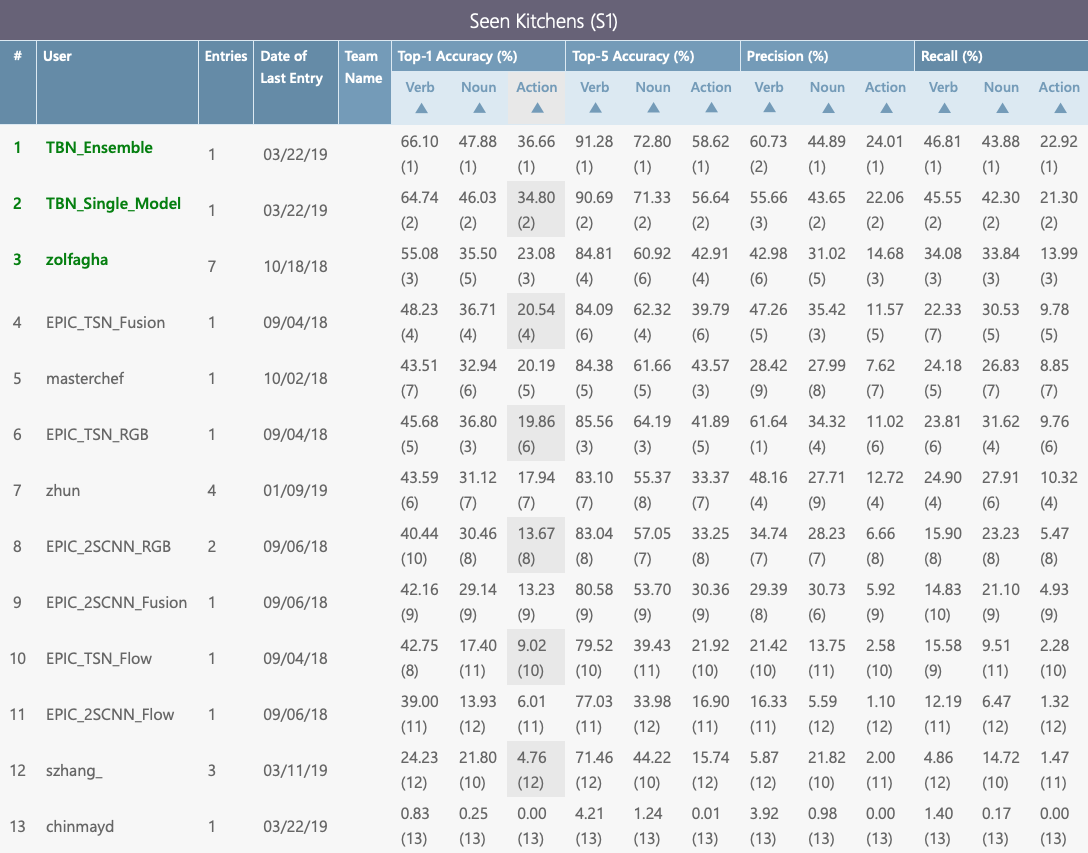}
 \includegraphics[width=\linewidth]{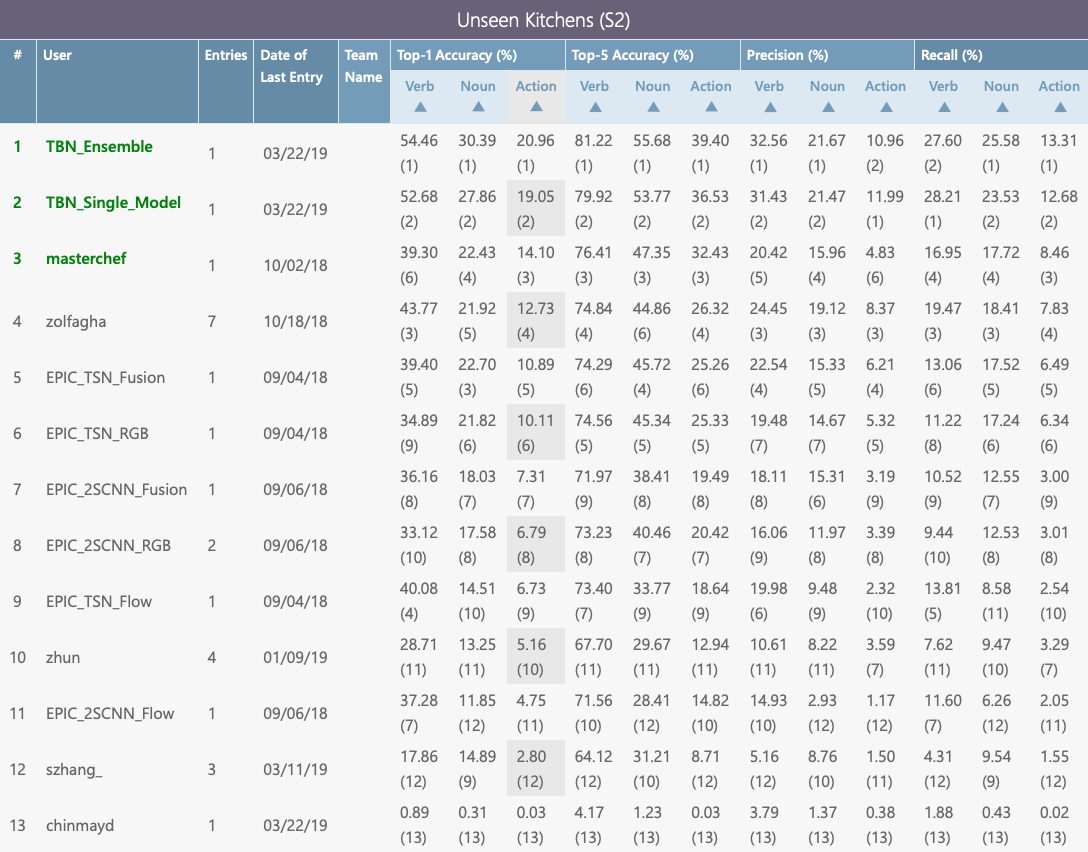}
 \caption{Our submission on the action recognition challenge for Seen (top) and Unseen (bottom) Kitchens.}
 \label{fig:challengeMar}
 \end{figure}

In Fig.~\ref{fig:challengeMar}, we show results for \textbf{Ours (TBN, Single Model)} and \textbf{Ours (TBN, Ensemble)}, as they appeared on the public leaderboard of the EPIC-Kitchens - Action recognition challenge on CodaLab at the time of submission (March 22nd 2019).
As noted in the paper, the single model TBN outperforms all other submissions by a clear margin, on both test sets \textbf{S1} and \textbf{S2}, and the results are further improved using an ensemble of TBNs trained with different TBW widths. 

As the challenge concluded, our model (TBN\_Ensemble) is ranked 2nd in the leaderboard. A snapshot of the leaderboard for the 2019 challenge is available at\\ \url{https://epic-kitchens.github.io/2019#results}.

\section{Per-class Multi-modal Fusion Results}
\label{sec:perclass}

\begin{figure*}[t]
	\centering
	\includegraphics[width=0.32\textwidth]{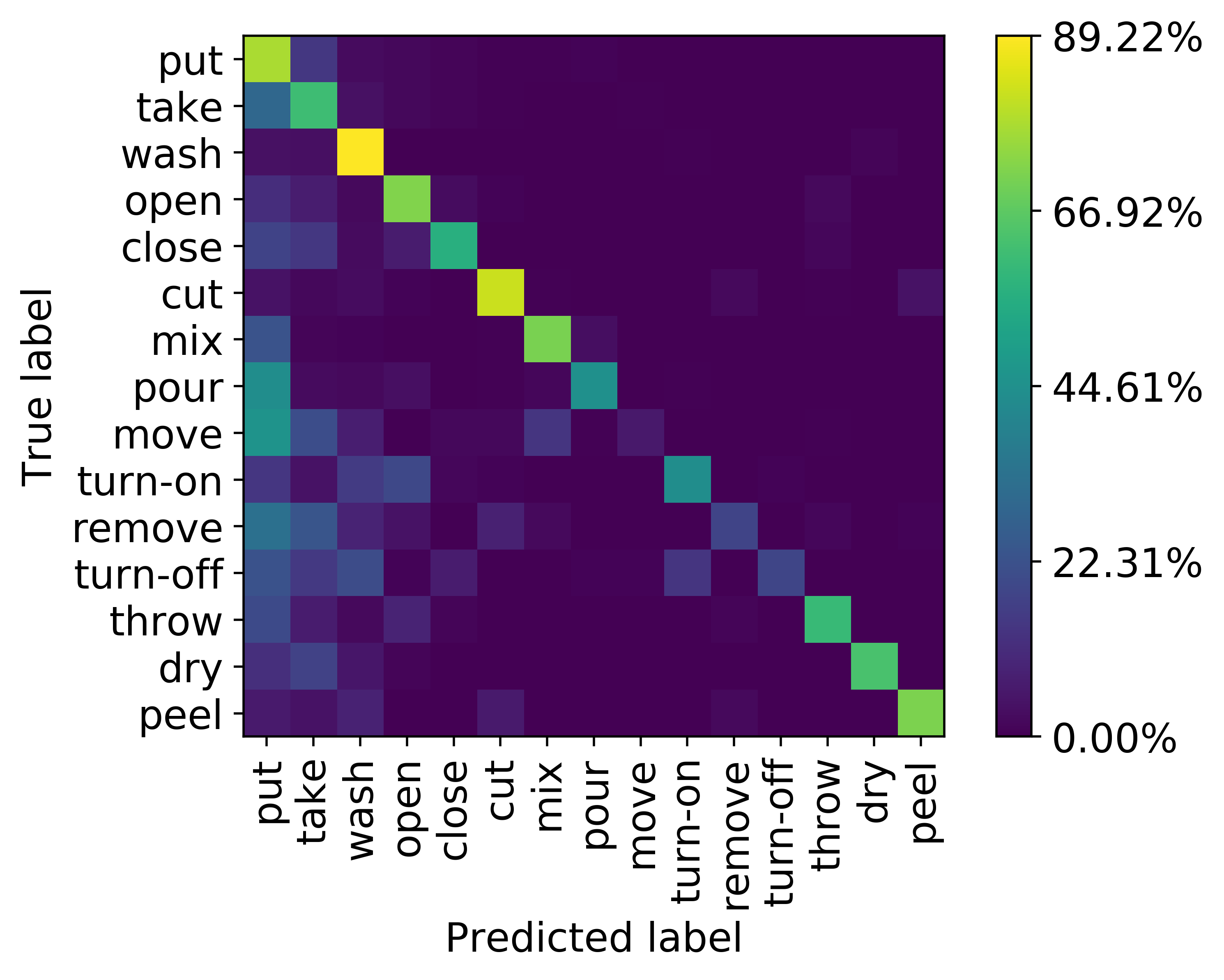}
	\includegraphics[width=0.32\textwidth]{figures/seen_verb_audio.png}
	\includegraphics[width=0.32\textwidth]{figures/seen_verb_diff.png} 
	\includegraphics[width=0.32\textwidth]{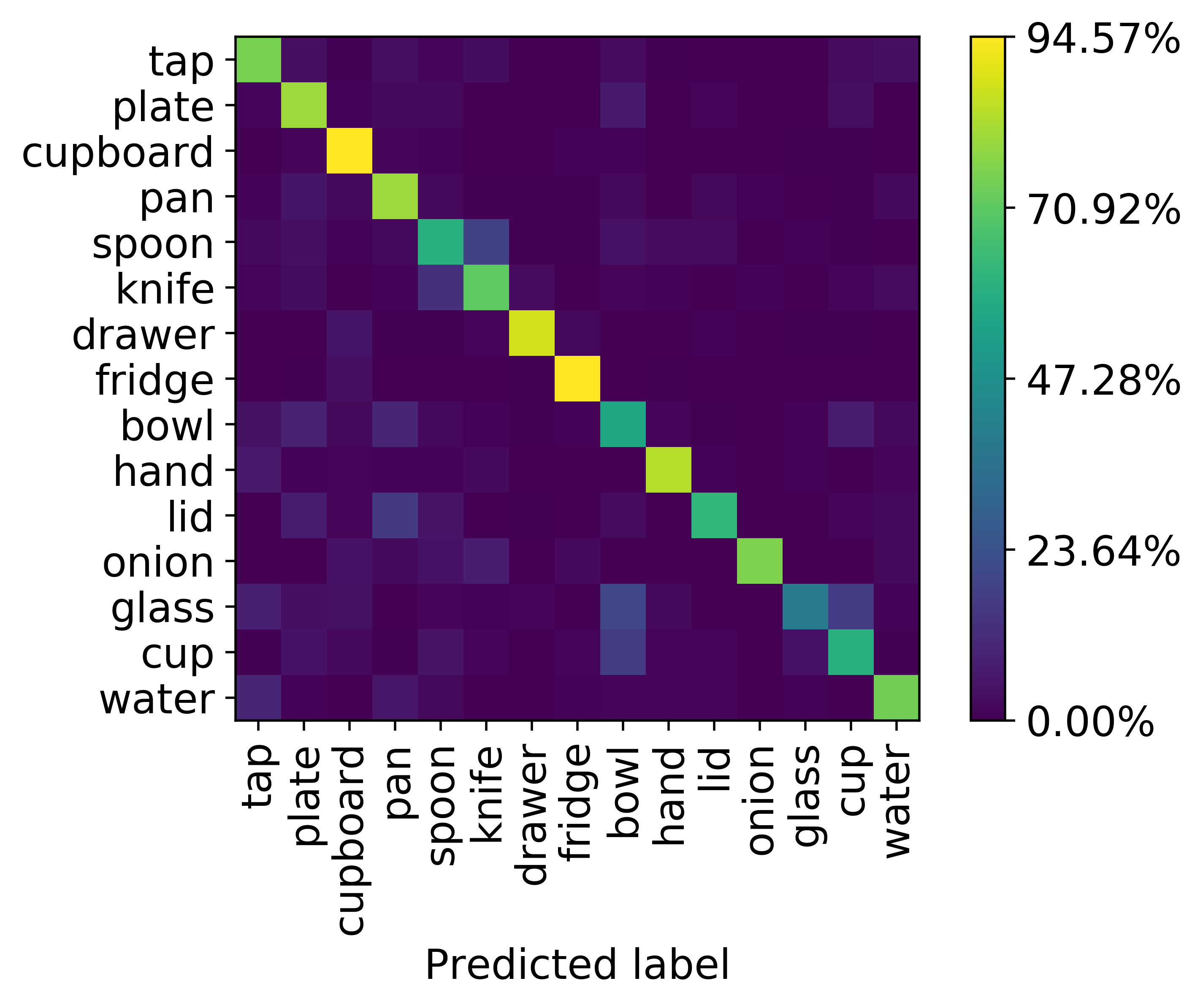}
	\includegraphics[width=0.32\textwidth]{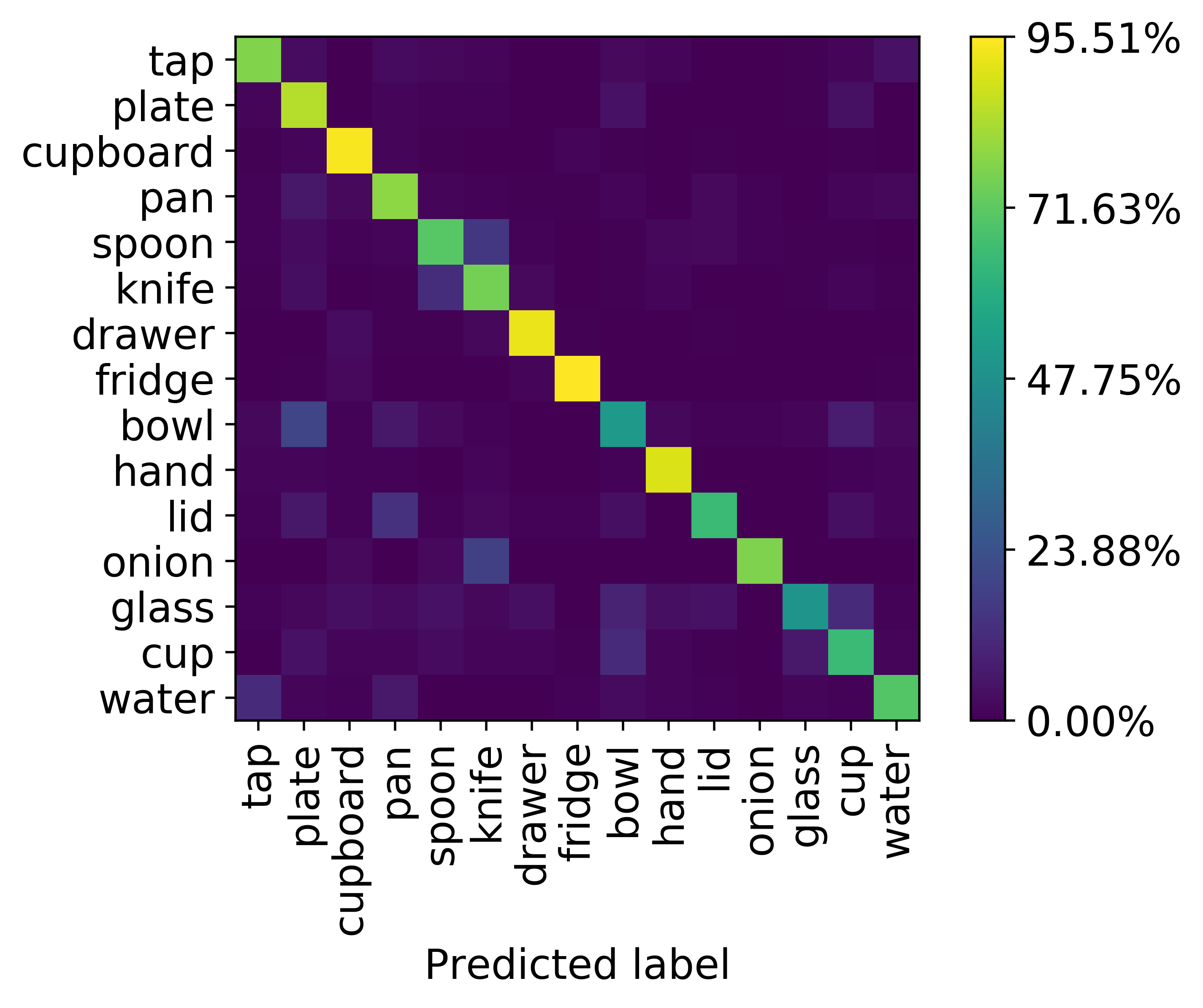}
	\includegraphics[width=0.32\textwidth]{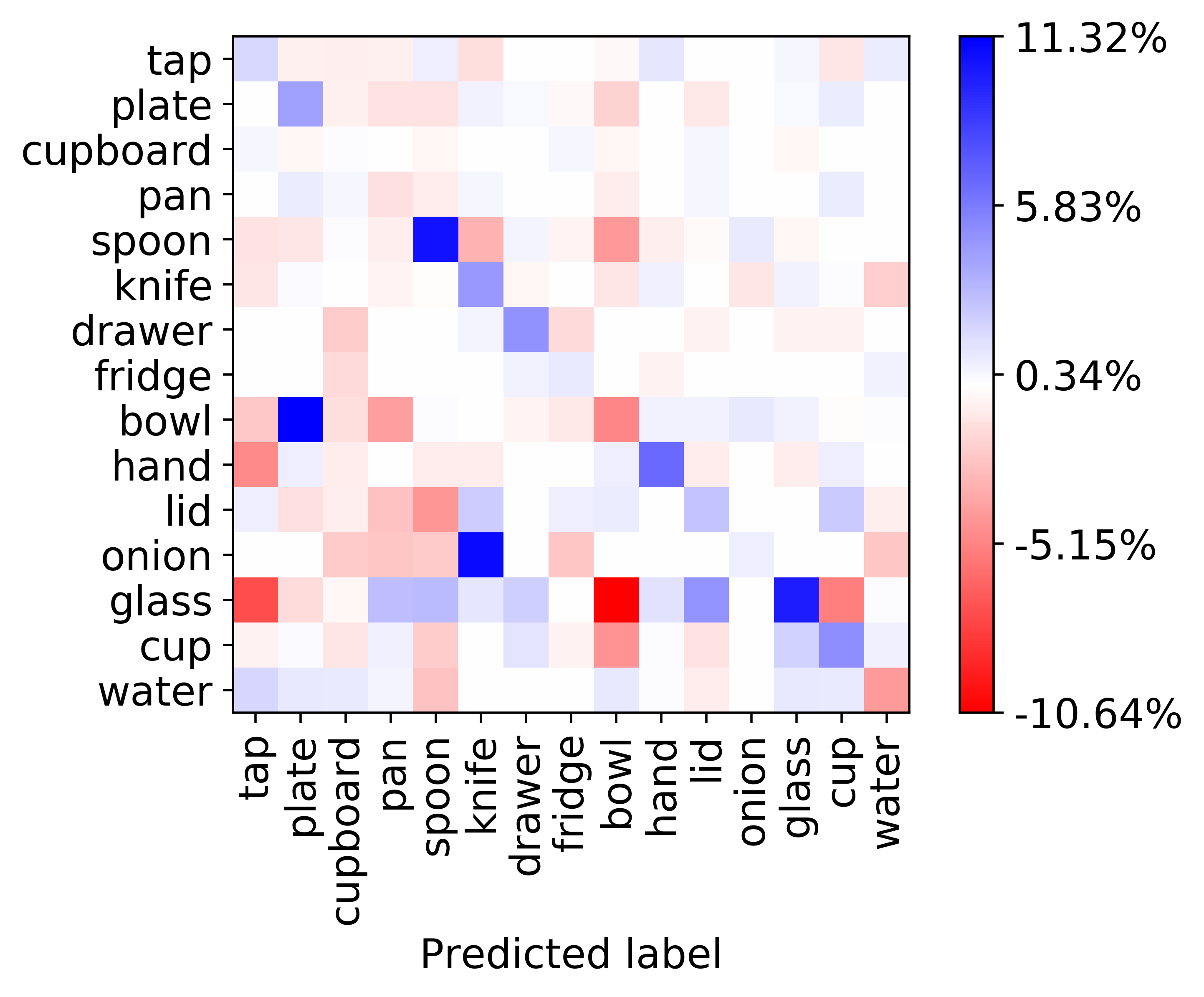} 
	\caption{Confusion matrices for largest-15 verb classes (top) and the largest-15 noun classes (bottom), without (left) and with (middle) audio, as well as their difference (right).}
	\label{fig:conf_mat1}
\end{figure*}
A complete version of Fig~\ref{fig:conf_mat} is available in Fig~\ref{fig:conf_mat1}. It shows the confusion matrices without and with the utilisation of audio for
the largest-15 verb and noun classes (in \textbf{S1}). The first confusion matrix show TBN (RGB+Flow), and the second shows TBN (RGB+Flow+Audio). Studying the difference (Fig~\ref{fig:conf_mat}~right) clearly demonstrates an increase~(blue) in confidence along the diagonal, and a decrease~(red) in confusion
elsewhere.

Table~\ref{tab:tail} shows a comparison of the performance of the top largest classes against the less represented classes, for individual modalities, and our proposed TBN. The classes are ranked by the number of examples in training, and the results are reported separately for the top-$10\%$ classes versus the rest which we refer to as \textit{tail classes}. The effect of fusion is more evident on the tail classes, $63\%$ improvement on tail vs. $34\%$ improvement on top-$10\%$ for verbs, and $50\%$ improvement on tail vs. $15\%$ improvement on top-$10\%$ for nouns. This finding shows that fusion in TBN decreases the effect of the class-imbalance. Furthermore, it is important to note that audio outperforms RGB and flow on
the tail verbs.

\begin{table}[t!]
    \centering
    {
    \begin{tabular}{ll|cccc}
        &  &  RGB & Flow & Audio & Fusion\\
         \hline
        \multirow{2}{*}{\rotatebox{90}{Verb}} & Top-10\% & 41.90 &  49.29 & 35.02 & 63.00\\
         & Tail  & 02.85 & 02.26 & 04.21 & 11.00\\
         \hline
        \multirow{2}{*}{\rotatebox{90}{Noun}} & Top-10\% &  39.52& 34.23  &22.36  & 46.50\\
         & Tail   & 06.22 & 02.38 & 04.76 & 13.02\\

    \end{tabular}}
    \caption{Comparison of mean class accuracy for the top 10\% of classes when ranked by class size 
    and the remaining~(tail) classes. It can be seen that fusion clearly has a greater effect on the tail classes, where single modalities perform very poorly.}
    \label{tab:tail}
\end{table}

\begin{table}[t!]
    \centering
    {\begin{tabular}{ll|c|c|c|c}
         & &  RGB & Flow & Audio & TBN\\
         \hline
        \rowcolor{gray}
         & open & 63.32 & 66.81 & 51.05 & 79.08\\
         \rowcolor{gray}
         & walk & 55.56 & 11.11 & 55.56 & 88.89\\
        \rowcolor{gray}
         & turn-on & 13.79 & 13.79 & 33.79 & 53.10\\
         \rowcolor{gray}
         & scoop & 02.27 & 04.55 & 02.27 & 18.18\\
         \rowcolor{gray}
         & look & 14.29 & 14.29 & 00.00 & 28.57\\
         \rowcolor{mediumgray}
         & scrape & 25.00 & 00.00 & 16.67 & 25.00\\
         \rowcolor{mediumgray}
         & hold & 00.00 & 20.00 & 00.00 & 20.00\\
         \rowcolor{mediumgray}
         & set & 33.33 & 00.00 & 16.67 & 33.33\\
         \rowcolor{mediumgray}
         & cook & 28.57 & 00.00 & 14.29 & 28.57\\
         \rowcolor{mediumgray}
         & finish & 00.00 & 00.00 & 16.67 & 16.67\\
         \rowcolor{lightgray}
         & insert & 01.79 & 00.00 & 07.14 & 03.57\\
         \rowcolor{lightgray}
         & divide & 00.00 & 40.00 & 00.00 & 20.00\\
         \rowcolor{lightgray}
         & sprinkle & 00.00 & 00.00 & 11.11 & 00.00\\
         \rowcolor{lightgray}
         & sample & 00.00 & 00.00 & 07.14 & 00.00\\
         \rowcolor{lightgray}
         & pat & 25.00 & 33.33 & 00.00 & 16.67
    \end{tabular}}
    \caption{Examples of predicted \textbf{verbs} on \textbf{S1} for each modality individually, and for TBN (Single Model).}
    \label{tab:verbs}
\end{table}

\begin{table}[t!]
    \centering
    {
    \begin{tabular}{ll|c|c|c|c}
         & &  RGB & Flow & Audio & TBN\\
         \hline
        \rowcolor{gray}
         & paella & 25.00 & 00.00 & 00.00 & 50.00\\
         \rowcolor{gray}
         & fridge & 83.25 & 80.10 & 60.73 & 87.96\\
         \rowcolor{gray}
         & hand & 56.38 & 59.73 & 43.62 & 76.51\\
         \rowcolor{gray}
         & sponge & 25.27 & 32.97 & 23.08 & 48.35\\
         \rowcolor{gray}
         & salt & 40.98 & 27.87 & 16.39 & 62.30\\
         \rowcolor{mediumgray}
         & switch & 50.00 & 00.00 & 75.00 & 75.00\\
         \rowcolor{mediumgray}
         & knife & 36.29 & 52.12 & 27.80 & 52.12\\
         \rowcolor{mediumgray}
         & salad & 14.29 & 19.05 & 04.76 & 19.05\\
         \rowcolor{mediumgray}
         & tortilla & 42.86 & 00.00 & 14.29 & 42.86\\
         \rowcolor{mediumgray}
         & leaf & 00.00 & 10.00 & 10.00 & 10.00\\
         \rowcolor{lightgray}
         & pizza & 100.00 & 09.09 & 36.36 & 72.73\\
         \rowcolor{lightgray}
         & fish & 90.00 & 00.00 & 00.00 & 50.00\\
         \rowcolor{lightgray}
        \rowcolor{lightgray}
         & bowl & 51.49 & 29.79 & 19.57 & 42.98\\
        \rowcolor{lightgray}
         & chicken & 31.58 & 15.79 & 07.89 & 26.32\\
         \rowcolor{lightgray}
         & paper & 03.70 & 00.00 & 14.81 & 07.41
    \end{tabular}}
    \caption{Examples of predicted \textbf{nouns} on \textbf{S1} for each modality individually, and for TBN (Single Model).}
    \label{tab:nouns}
\end{table}
 
 In Tables~\ref{tab:verbs} and~\ref{tab:nouns}, we show per-class accuracies on~\textbf{S1}, on selected verbs and nouns, respectively. We arrange the chosen set of verbs and nouns in three main categories: \textbf{top:} TBN outperforms the best individual modality, \textbf{mid:}~TBN performs comparably with the best modality, and \textbf{bottom:}~TBN performs worse than the best individual modality.  We shade the rows reflecting these three groups in the order mentioned above. 

A few conclusions could be made from these tables about the advantages of the proposed mid-level fusion:
\begin{enumerate}
\item Fusion can improve results when all modalities are individually performing well for both verb and noun classes (e.g. `open', `fridge'), as well as when all modalities are under-performing (e.g. `scoop', `salad').
\item Fusion can though be difficult at times, particularly when two of the three modalities are uninformative (e.g.~`divide', `fish').
\item All nouns for which audio is outperforming other modalities have distinct sounds (e.g.~`switch', `paper').
\item Similarly, audio is least distinctive when the noun does not have a sound per se or its sound depends on the action (e.g.~`chicken', `salt').
\end{enumerate}

\section{Code and Models}
Python code of our TBN model, and pre-trained model on EPIC-Kitchens is available at\\ \url{{http://github.com/ekazakos/temporal-binding-network}}

\end{document}